\documentclass[conference]{IEEEtran}
\IEEEoverridecommandlockouts
\usepackage{cite}
\usepackage{amsmath,amssymb,amsfonts}
\usepackage{algorithmic}
\usepackage{graphicx}
\usepackage{textcomp}
\usepackage{xcolor}
\def\BibTeX{{\rm B\kern-.05em{\sc i\kern-.025em b}\kern-.08em
    T\kern-.1667em\lower.7ex\hbox{E}\kern-.125emX}}
\begin{document}

\title{Balanced k-Means Clustering on an Adiabatic Quantum Computer\\
\thanks{This manuscript has been authored by UT-Battelle, LLC under Contract No. DE-AC05-00OR22725 with the U.S. Department of Energy.  The United States Government retains and the publisher, by accepting the article for publication, acknowledges that the United States Government retains a non-exclusive, paid-up, irrevocable, world-wide license to publish or reproduce the published form of this manuscript, or allow others to do so, for United States Government purposes.  The Department of Energy will provide public access to these results of federally sponsored research in accordance with the DOE Public Access Plan (http://energy.gov/downloads/doe-public-access-plan).}
}

\author{\IEEEauthorblockN{Davis Arthur}
\IEEEauthorblockA{\textit{Auburn University} \\
Alabama, USA \\
dsa0013@auburn.edu}
\and
\IEEEauthorblockN{Prasanna Date}
\IEEEauthorblockA{\textit{Oak Ridge National Laboratory} \\
Tennessee, USA \\
datepa@ornl.gov}
}

\maketitle

\begin{abstract}
Adiabatic quantum computers are a promising platform for approximately solving challenging optimization problems. We present a quantum approach to solving the balanced $k$-means clustering training problem on the D-Wave 2000Q adiabatic quantum computer. Existing classical approaches scale poorly for large datasets and only guarantee a locally optimal solution. We show that our quantum approach better targets the global solution of the training problem, while achieving better theoretic scalability on large datasets. We test our quantum approach on a number of small problems, and observe clustering performance similar to the best classical algorithms.
\end{abstract}


\section{Introduction}
\label{intro}
\quad Applications of machine learning are prevalent throughout the modern world. While their tasks vary greatly in purpose and scale, all machine learning models must be trained before they can be deployed for practical use. In some cases, the training process is extremely time consuming, even on the most powerful classical computers. This is particularly true for models with NP-hard or NP-complete training problems such as $k$-means clustering \cite{Aloise2009np}, neural networks \cite{Blum1988np}, decision tree learning \cite{Hyafil1976np}, etc. 

Quantum computers offer an alternative platform for efficiently solving computationally challenging problems. For instance, the D-Wave 2000Q adiabatic quantum computer approximately solves the NP-complete quadratic unconstrained binary optimization (QUBO) problem efficiently. The D-Wave quantum computer has already been used for a number of machine learning tasks including training a support vector machine \cite{Willsch2020svm}, training a restricted Boltzmann machine \cite{dixit2020rbm} \cite{date2019classical}, linear regression \cite{date2020adiabatic} and matrix factorization for feature learning \cite{O_Malley2018featurelearning}. While modern quantum computers are too small and error-prone to effectively solve large problems, their scale and fidelity are expected to improve dramatically in time \cite{Preskill2018nisq}. 

In this paper, we use the D-Wave 2000Q adiabatic quantum computer to perform a special case of $k$-means clustering. $k$-means clustering is a popular machine learning model that partitions a set of $N$ data points into $k$ clusters such that each cluster is made up of similar points. Similarity is measured by the statistical variance within each cluster. We focus on balanced $k$-means clustering, which requires that each cluster contains approximately the same number of points. Balanced clustering models are used in a variety of domains including network design \cite{gupta2003networks}, marketing \cite{Ghosh2005marketing}, and document clustering \cite{Banerjee2003documentclustering}. 

Classically, it is computationally challenging to find the exact solution to the balanced $k$-means training problem. Thus, existing algorithms converge after finding a locally optimal solution. In the worst case, this can still require large computational resources, especially as problem size scales. Due to these challenges, we explore the prospect of training the balanced $k$-means model on an adiabatic quantum computer. First, we outline a QUBO formulation of the balanced $k$-means clustering training problem. We then theoretically analyze our formulation, comparing our quantum approach to current classical algorithms. Next, we empirically analyze the clustering performance and scalability of our quantum approach on synthetic classification data sets. Finally, we analyze the clustering performance of our approach on portions of the Iris benchmark data set.

\section{Related Work}

\quad The $k$-means clustering model is one of the most widely used unsupervised machine learning techniques. Classically, the model is usually trained through an iterative approach known as Lloyd's algorithm. Hartigan and Wong show that the time complexity of this approach is $\mathcal{O}(Nkdi)$ where $N$ is the number of data points, $k$ is the number of clusters, $d$ is the dimension of the data set, and $i$ is the number of iterations before the algorithm converges \cite{Hartigan1979KMeans}. Arthur and Vassilvitskii prove that for random cluster initialization, $i = 2^{\Omega(\sqrt{n})}$ with high probability \cite{Arthur2006worstcasei}. Therefore, Lloyd's algorithm has superpolynomial time complexity.

Many different implementations and variations of Lloyd's algorithm have been proposed to avoid long training times or poor clustering performance. Na et al. propose an efficient implementation that reduces the number of required distance calculations without compromising clustering quality \cite{Na2010efficient}. Celebi et al. compare the impact of several different centroid initialization methods on clustering performance and run time \cite{Celebi2012initialization}. Kapoor and Singhal observe a reduction in run time and superior clustering results when sorting input data before training the $k$-means model \cite{Kapoor2017sorted}. The Scikit-learn implementation of Lloyd's algorithm bounds the number of iterations by a constant, effectively reducing the time complexity to $\mathcal{O}(Nkd)$ \cite{Pedregosa2011scikit}. We have used this implementation as a point of comparison to our quantum approach.

Constrained $k$-means models, such as balanced $k$-means clustering, are common in applications where additional knowledge regarding the training data or the form of a plausible solution is known. Sometimes constrained $k$-means models are also used in instances where the generic $k$-means algorithm is likely to converge to a suboptimal solution \cite{bradley2000constrained}. Bradley et al. propose an algorithm that enforces a minimum bound on cluster size \cite{bradley2000constrained}. This approach reduces to balanced clustering when the minimum cluster size is $\operatorname{floor}(N / k)$. Ganganath et al. present a constrained $k$-means clustering algorithm in which the size of each cluster is specified prior to training the model \cite{Ganganath2014samesizekclustering}. Malinen et al. propose an efficient balanced $k$-means clustering algorithm that runs in $\mathcal{O}(N^3)$ time \cite{Malinen2014balancedcomplexity}. This algorithm will be used as a point of comparison to our quantum approach. 

Quantum approaches to training clustering models have been proposed as well. Khan et al. implement a quantum algorithm similar to Lloyd's algorithm on the IBMQX2 universal quantum computer \cite{khan2019kmeans}. Ushijima-Mwesigwa et al. demonstrate partitioning a graph into $k$ parts concurrently using quantum annealing on the D-Wave 2X machine \cite{UshijimaMwesigwa2017partitioning}. Neukart et al. propose a quantum-classical hybrid approach to clustering \cite{neukart2018quantumassisted}. Wereszczynski et al. demonstrate the performance of a novel quantum clustering algorithm on small data sets using the D-Wave 2000Q \cite{weresz2018clustering}. Bauckhage et al. propose a QUBO formulations to binary clustering ($k = 2$) \cite{bauckhage2018k2clustering} and $k$-medoids clustering \cite{bauckhage2019qubo}. Kumar et al. present a QUBO formulation for $k$-clustering that approximates the $k$-means model \cite{Kumar2018combinatorialclustering}. We have previously formulated three machine learning problems as QUBO problems \cite{date2020qubo}.

While many quantum clustering algorithms have been proposed, none target the exact solution to the $k$-means or balanced $k$-means clustering model. Instead, they are heuristic approaches that approximate the $k$-means optimization problem. We propose a QUBO formulation that is identical to the balanced $k$-means training problem. We also tested our approach on both synthetic and benchmark data.

\section{QUBO Formulation}
\quad Adiabatic quantum computers are able to find the global minimum of the quadratic unconstrained binary optimization (QUBO) problem, which can be stated as follows:
\begin{align}
    \min_{z \in \mathbb{B}^M} z^T A z \label{eq:qubo}
\end{align}
where 
$\mathbb{B} = \{0, 1\}$ is the set of binary numbers,
$z \in \mathbb{B}^M$ is the binary decision vector,
and $A \in \mathbb{R}^{M \times M}$ is the real-valued $M \times M$ QUBO matrix. Our goal is to convert the balanced $k$-means training problem into this form.

The $k$-means clustering model, aims to partition a data set $X = \{ x_1, x_2, ..., x_N \}$ into $k$ clusters $\Phi = \{\phi_1, \phi_2, ..., \phi_k\}$. The centroid of cluster $\phi_i$ is denoted as $\mu_i$. Formally, training the $k$-means clustering model is expressed as:
\begin{align}
    \min_{\Phi} \sum_{i = 1}^k \sum_{x \in \phi_i} ||x - \mu_i||^2 \label{eq:kclusters1}
\end{align}
Utilizing the law of total variance, the training problem can be rewritten as:
\begin{align}
    \min_{\Phi} \sum_{i = 1}^k \frac{1}{2|\phi_i|} \sum_{x, y \in \phi_i} || x - y ||^2 \label{eq:kclusters2}
\end{align}
In the case that each cluster is of equal size (i.e. balanced), $|\phi_i|$ is constant, and Problem \ref{eq:kclusters2} reduces to:
\begin{align}
    \min_{\Phi} \sum_{i = 1}^k \sum_{x, y \in \phi_i} || x - y ||^2
    \label{eq:kmeanssamesize}
\end{align}
To formulate Problem \ref{eq:kmeanssamesize} as a QUBO problem, it will be useful to define a matrix $D \in \mathbb{R}^{N \times N}$ where each element is given by $d_{ij} = ||x_i - x_j||^2$.
We also define a binary matrix $\hat{W} \in \mathbb{B}^{N \times k}$ such that $\hat{w}_{ij} = 1$ if and only if point $x_i$ belongs to cluster $\phi_j$. This use of binary variables is identical to the ``one-hot encoding" quantum clustering method proposed by Kumar et al. \cite{Kumar2018combinatorialclustering}. Since we are assuming clusters of the same size, each column in $\hat{W}$ should have approximately $N / k$ entries equal to 1. Additionally, since each data point belongs to exactly one cluster, each row in $\hat{W}$ must contain exactly one entry equal to 1. Using this notation, the inner sum in Problem \ref{eq:kmeanssamesize} can be rewritten:
\begin{align}
    \sum_{x, y \in \phi_j} || x - y ||^2 = \hat{w}{'}_j^T D \hat{w}'_j
    \label{eq:sumrelation}
\end{align}
where $\hat{w}'_j$ is the $j^\text{th}$ column in $\hat{W}$. From this relation, we can cast Problem \ref{eq:kmeanssamesize} into a constrained binary optimization problem.
First, we vertically stack the $Nk$ binary variables in $\hat{W}$ as follows:
\begin{align}
    \hat{w} = 
    [\hat{w}_{11} \ldots \hat{w}_{N1} \ \hat{w}_{12} \ldots \hat{w}_{N2} \ldots \hat{w}_{1k} \ldots \hat{w}_{Nk}]^T \label{eq:kmeansz}
\end{align}
Provided the constraints on $\hat{w}$ are upheld, Problem \ref{eq:kmeanssamesize} is equivalent to:
\begin{align}
    \min_{\hat{w}} \hat{w}^T (I_k \otimes D) \hat{w}
    \label{eq:kmeansconstrained}
\end{align}
where $I_k$ is the $k$-dimensional identity matrix.

We can remove the constraints on $\hat{w}$ by including penalty terms that are minimized when all conditions are satisfied. First, we account for the constraint that each cluster must contain approximately $N/k$ points. For a given column $\hat{w}'_j$ in $\hat{W}$, this can be enforced by including a penalty of the form:
\begin{align}
    \alpha (\hat{w}{'}_j^T \hat{w}'_j - N/k)^2 
    \label{eq:krow1}
\end{align}
where $\alpha$ is a constant factor intended to make the penalty large enough that the constraint is always upheld. Dropping the constant term $\alpha(N/k)^2$, this penalty is equivalent to $\hat{w}{'}_j^T \alpha F \hat{w}'_j$ where $F$ is defined as:
\begin{align}
    F = 1_N - \frac{2N}{k} I_N
\end{align}
In the expression above, $1_N$ refers to an $N \times N$ matrix where each element is equal to 1. Using this formulation, the sum of all column constraint penalties is:
\begin{align}
    \hat{w}^T (I_k \otimes \alpha F ) \hat{w} \label{eq:kmeanscolumns}
\end{align}

Next, we account for the constraint that each point belongs to exactly $1$ cluster. For a given row $\hat{w}_i$, this can be enforced by including a penalty of the form:
\begin{align}
    \beta (\hat{w}_i^T \hat{w}_i - 1)^2
\end{align}
where $\beta$ is a constant with the same purpose as $\alpha$ in Equation \ref{eq:krow1}. Dropping the constant term, this penalty is equivalent to $\hat{w}_i^T \beta G \hat{w}_i$ where $G$ is defined as:
\begin{align}
    G = 1_k - 2 I_k
\end{align} 
To find the sum of all row constraint penalties, we first convert the binary vector $\hat{w}$ into the form $\hat{v}$ shown below:
\begin{align}
    \hat{v} = [w_{11} \ldots w_{1k} \ w_{21} \ldots w_{2k} \ldots w_{N1} \ldots w_{Nk}]^T
\end{align}
This can be accomplished through a linear transformation $Q \hat{w}$ where each element in $Q \in \mathbb{B}^{Nk \times Nk}$ is defined as:
\begin{align}
    q_{ij} = 
    \begin{cases} 
      1 & j =  N \operatorname{mod}(i - 1, k) + \lfloor \frac{i - 1}{k} \rfloor + 1 \\
      0 & \text{else} \\
   \end{cases}
\end{align}
After the transformation, the sum of all row constraint penalties is given by $\hat{v}^T (I_N \otimes \beta G) \hat{v}$. This sum can be equivalently expressed as:
\begin{align}
    \hat{w}^T Q^T (I_N \otimes \beta G) Q \hat{w} \label{eq:kmeansrow}
\end{align}
Combining the column and row penalties with the constrained binary optimization problem from Equation \ref{eq:kmeansconstrained}, Problem \ref{eq:kmeanssamesize} can be rewritten as:
\begin{align}
    \min_{\hat{w}} \hat{w}^T (I_k \otimes (D + \alpha F) + Q^T (I_N \otimes \beta G) Q) \hat{w} \label{eq:finalkmeans}
\end{align}
This is identical to Equation \ref{eq:qubo} with $z = \hat{w}$ and $A = (I_k \otimes (D + \alpha F) + Q^T (I_N \otimes \beta G) Q)$. Thus, we have converted the balanced $k$-means training problem (Equation \ref{eq:kmeanssamesize}) into a QUBO problem which can be solved on adiabatic quantum computers. Provided $N$ is divisible by $k$, and $\alpha$ and $\beta$ are large enough to ensure all constraints are upheld, Problem \ref{eq:finalkmeans} and Problem \ref{eq:kmeanssamesize} share the same global solution. 

\subsection{Implementation Details}

\quad In order to achieve good performance on quantum hardware, $\alpha$ and $\beta$ must be chosen such that the penalty for violating a constraint is large, but not so large as to overshadow the importance of minimizing within cluster variance. In practice we achieved the best performance when defining $\alpha$ and $\beta$ as follows:
\begin{align}
    \alpha = \frac{\operatorname{max}(D)}{2(N / k) - 1} \\
    \beta = \operatorname{max}(D)
\end{align}
where $\operatorname{max}(D)$ is the maximum element in $D$. 

By choosing these values, we scale $F$ and $G$ such that the maximum value in each scaled matrix is equal to the maximum value in $D$. Assuming the training data set has well defined clusters, the maximum element of $D$ is much larger than the average squared distance between two points in a given cluster. Therefore, these multipliers assure that the penalty for violating a constraint is almost always larger than the penalty for a poor clustering assignment. By multiplying $F$ by a smaller factor than $G$, we also guarantee that row constraints are more strictly enforced than column constraints. This is desirable since we would like to permit small violations of the equal-size cluster constraint when $N$ is not divisible by $k$.

In practice the quantum annealing process is not perfect, and instances occur in which a point is assigned to multiple clusters or not assigned to any cluster at all. If quantum annealing assigns a point to multiple clusters, we consider the point to belong to the cluster with the smaller index. If quantum annealing does not assign a point to any cluster, we consider the point to belong to the first cluster.

\section{Results and Analysis}

\begin{figure*}[t!]
\centering
  \includegraphics[scale = 0.35]{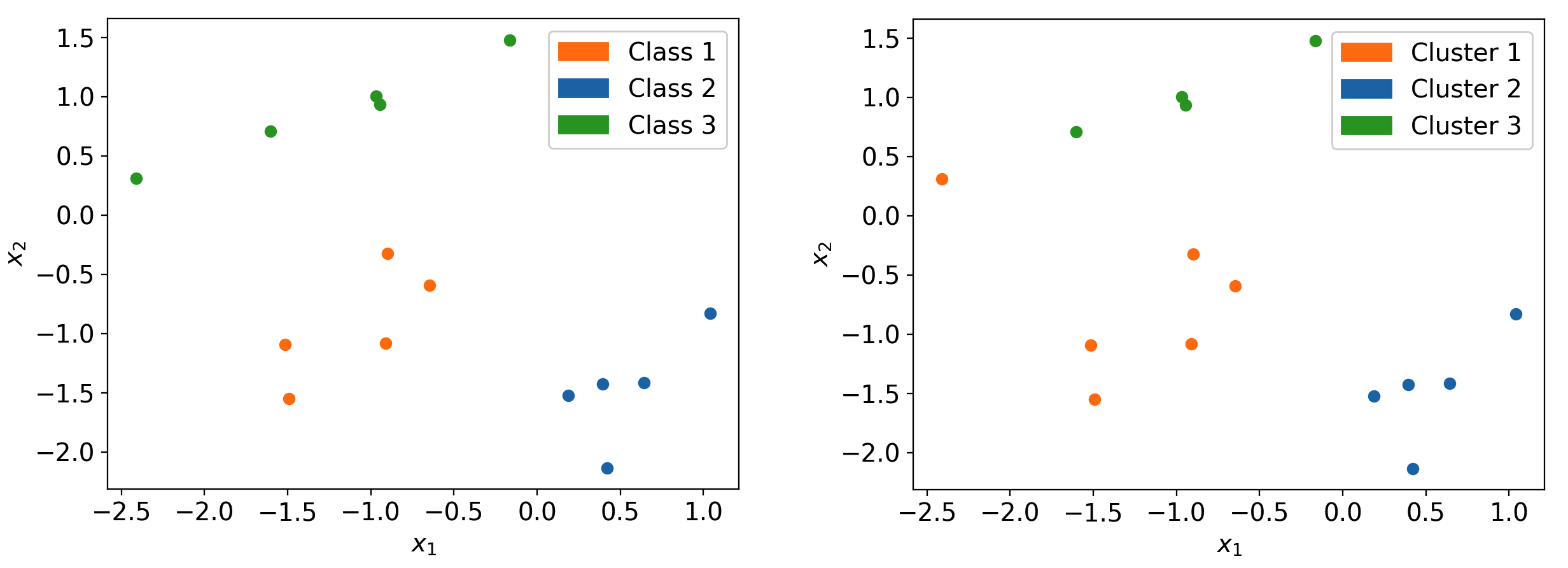}
\caption{Example of a synthetic data set containing $N = 15$ points partitioned into 3 classes (left). The quantum algorithm correctly partitioned all but one point (right).}
\label{fig:demo}   
\end{figure*}

\subsection{Theoretical Analysis}
\label{sub:theory}
\quad The generic $k$-means clustering problem stated in Equation \ref{eq:kclusters1} and the balanced $k$-means clustering problem stated in Equation \ref{eq:kmeanssamesize} both contain $\mathcal{O}(Nd)$ data and $\mathcal{O}(N)$ variables (where each variable indicates the cluster assignment of a given data point). In our QUBO formulation of balanced $k$-means clustering, we introduce $k$ binary variables for each variable in the original problem. Thus, the total number of variables in Equation \ref{eq:finalkmeans} is $\mathcal{O}(Nk)$. This translates to a quadratic qubit footprint of $\mathcal{O}(N^2 k^2)$ using an efficient embedding algorithm such as \cite{date2019efficiently}.

It has been shown to require $\mathcal{O}(N^{kd + 1})$ time to exactly solve the generic $k$-means clustering problem (Problem \ref{eq:kclusters1}) \cite{Inaba1994kworstcase}. Alternatively, a locally optimal solution can be found in $\mathcal{O}(Nkdi)$ time using Lloyd's algorithm. The Scikit-learn approach to $k$-means is able to effectively reduce the time complexity to $\mathcal{O}(Nkd)$ by bounding the number of iterations by a constant and performing Lloyd's algorithm multiple times from different centroid initializations. While this approach cannot guarantee a locally optimal solution, it achieves high quality clustering performance in practice.

The time complexity required to exactly solve the balanced $k$-means clustering problem has not been thoroughly analyzed. However, a locally optimal solution to Problem \ref{eq:kmeanssamesize} can be found in $\mathcal{O}(N^3)$ time using the classical approach proposed by Malinen et al. \cite{Malinen2014balancedcomplexity}. To compare this to our quantum approach, we first determine the time complexity for converting Equation \ref{eq:kmeanssamesize} into a QUBO problem. To do so, we rewrite Equation \ref{eq:finalkmeans} as follows:
\begin{align}
    & \min_{W} \sum_{l = 1}^k \sum_{j = 1}^N \sum_{i = 1}^N  \sum_{m = 1}^d w_{il} (x_{im} - x_{jm})^2 w_{jl} 
    \nonumber \\ &
    + \alpha \sum_{l = 1}^k \sum_{j = 1}^N \sum_{i = 1}^N w_{il} f_{ij} w_{jl} 
    + \beta \sum_{l = 1}^N \sum_{j = 1}^k \sum_{i = 1}^k w_{li} g_{ij} w_{lj}
    \label{eq:kmeanscomplexity}
\end{align}
From Equation \ref{eq:kmeanscomplexity}, the worst case time complexity is $\mathcal{O}(N^2 k d)$, which is dominated by the first term. For practical purposes, solving the QUBO problem through quantum annealing can be done in constant time. Therefore, the total time complexity of the quantum algorithm is $\mathcal{O}(N^2 k d)$. Provided $kd < N$, this time complexity is better than the time complexity of the best classical balanced $k$-means clustering algorithm $(\mathcal{O}(N^3))$. However, it is worse than the Scikit-learn implementation of generic $k$-means clustering $(\mathcal{O}(Nkd))$.

\subsection{Empirical Analysis}
\label{sub:empirical}
\subsubsection{Methodology and Performance Metrics}
Our quantum approach was tested on the D-Wave 2000Q adiabatic quantum computer. We compare the performance of our approach to the Scikit-learn implementation of classical $k$-means as well as our own implementation of the classical balanced k-means algorithm with the best time complexity \cite{Malinen2014balancedcomplexity}. Note that the Scikit-learn implementation of $k$-means searches for a solution to Problem \ref{eq:kclusters1}, while the classical balanced $k$-means algorithm and our quantum approach search for a solution to Problem \ref{eq:kmeanssamesize}. The Scikit-learn algorithm is still a valid point of comparison since the solution to both problems should be very similar for all data sets used in our experiments.

We use two performance metrics to compare the three algorithms: (i) adjusted rand index and (ii) total computing time. In the quantum approach, total computing time is composed of the time required to convert the problem into a QUBO problem, the time required to embed the QUBO problem on the hardware, the time for the quantum computer to solve the QUBO problem (annealing time), and the time required to extract the clustering information from the binary solution (postprocessing time).

\subsubsection{Data Generation} \label{sub:synthetic} 
We tested our algorithm on synthetic classification data sets created using the \textit{make\_classification} function in the Scikit-learn datasets package. Each data set contains $N$ points, $k$ classes, $1$ cluster per class, and $d$ features. This function generates a data set where each cluster is centered at one of the vertices of a $d$-dimensional hypercube with side length $2.0$. The points are then generated from a normal distribution (standard deviation of $1.0$) about their cluster center. For all experiments, each class was made up of exactly $N/k$ points.

\subsubsection{Hardware Configuration} 
Preprocessing and postprocessing for our quantum approach and entire classical approach were run on a machine with 2.7 GHz Dual-Core Intel i5 processor and 8 GB 1,867 MHz DDR3 memory. The quantum approach also used the D-Wave 2000Q quantum computer, which had 2,048 qubits and about 5,600 inter-qubit connections. For all experiments, each quantum annealing operation is performed 100 times, and only the ground state is used. 

\begin{table*}[t!]
\centering
\caption{Number of binary variables and average number of qubits used in the quantum approach.}
\label{tab:numqubits}       
\begin{tabular}{c|ccccccccc}
\hline\noalign{\smallskip}
 ($N$, $k$) & (16, 2) & (24, 2) & (32, 2) & (12, 3) & (15, 3) & (21, 3) & (8, 4) & (12, 4) & (16, 4) \\
\noalign{\smallskip}\hline\noalign{\smallskip}
\shortstack{Variables} & 32 & 48 & 64 & 36 & 45 & 63 & 32 & 48 & 64 \\
\shortstack{Qubits} & 185 & 429 & 794 & 244 & 381 & 743 & 209 & 456 & 806 \\
\noalign{\smallskip}\hline
\end{tabular}
\end{table*}

\subsubsection{Adjusted Rand Index} 
The adjusted rand index (ARI) is a metric used to compare the similarity of two partitions of a data set. This metric ranges from -1 to 1, with larger values indicating that the two partitions are similar. We use the adjusted rand index to compare the ground truth labels of a classification data set to the partitioning produced by a clustering algorithm. A value of 1 indicates that the algorithm perfectly partitioned the data, and values close to 0 are reflective of random clustering.

If $\Phi = \{ \phi_1, \phi_2, ..., \phi_k \}$ is the partitioning produced by a given clustering algorithm, and $Y = \{ Y_1, Y_2, ... Y_k \}$ is the partitioning produced by the target function, the overlap of $\Phi$ and $Y$ is given in the contingency table $[n_{ij}]$ where $n_{ij} = |\phi_i \cap Y_j|$. We denote the sum over all entries in the $i^\text{th}$ row of the table as $a_i = \sum_{m=1}^k n_{im}$ and the sum over all entries in the $j^\text{th}$ column of the table as $b_j = \sum_{m=1}^k n_{mj}$. Using this notation, the adjusted rand index is defined below:

\begin{align}
    ARI = \frac{\sum_{ij} {n_{ij} \choose 2} - \left[ \sum_i {a_i \choose 2} \sum_j {b_j \choose 2} \right] / {N \choose 2}}
    {\frac{1}{2} \left[ \sum_i {a_i \choose 2} + \sum_j {b_j \choose 2}\right] - \left[ \sum_i {a_i \choose 2} \sum_j {b_j \choose 2} \right] / {N \choose 2}}
\end{align}

\subsubsection{Clustering Synthetic Data Sets}
\label{sec:synth}
\begin{figure}[t!]
\centering
  \includegraphics[scale = 0.54]{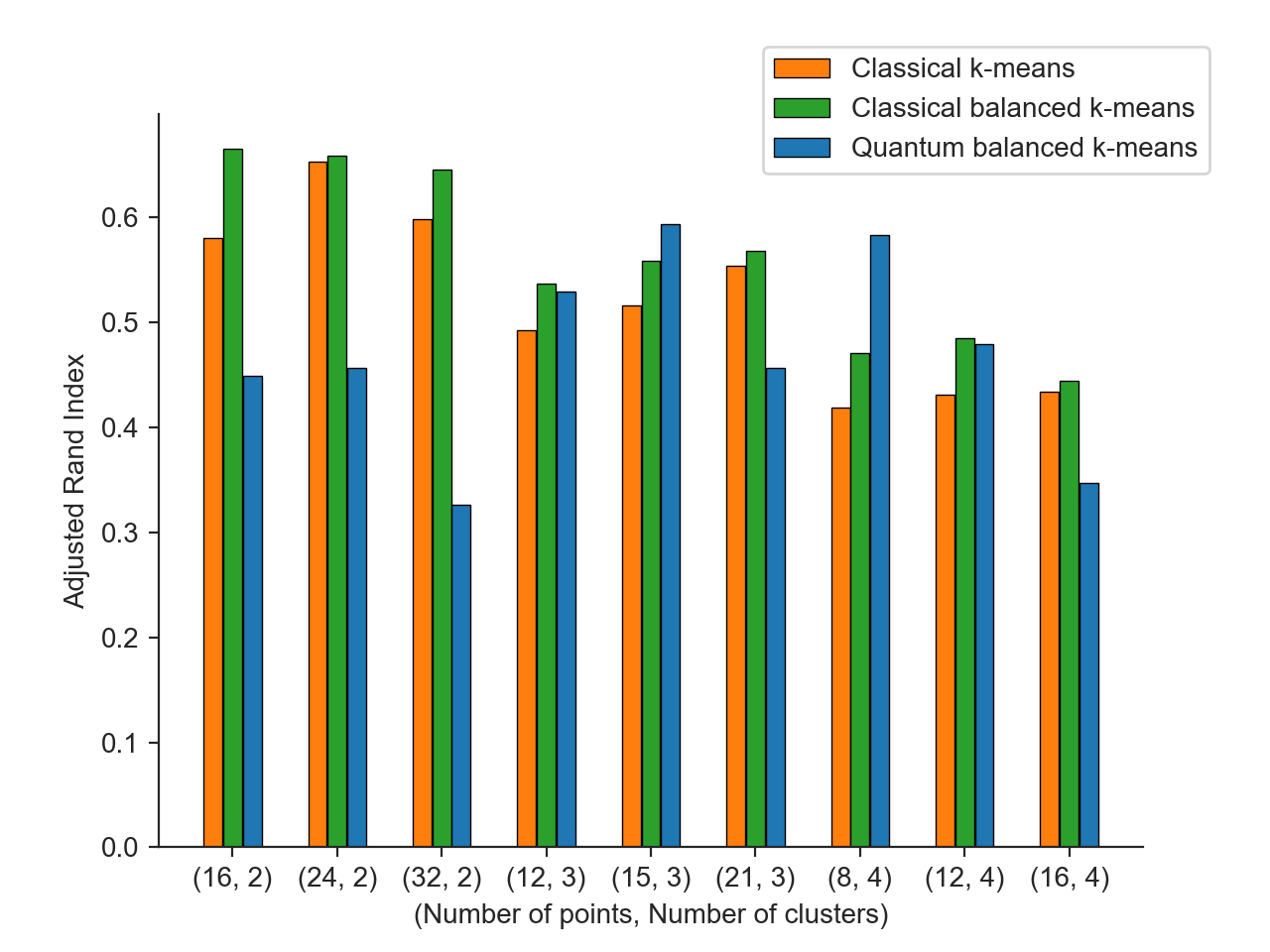}
\caption{Adjusted rand index of clustering solutions produced by classical k-means (orange bar), classical balanced k-means (green bar), and quantum balanced k-means (blue bar).}
\label{fig:randsynth}   
\end{figure}

We compare the clustering quality produced by classical $k$-means, classical balanced $k$-means, and quantum balanced k-means on a number of small synthetic data sets. For each problem type (defined by the number of points and number of clusters), all three algorithms were run on $50$ synthetic classification data sets. The average adjusted rand index of each clustering algorithm is reported in Figure \ref{fig:randsynth}.

For most experiments, the classical balanced k-means algorithm had the best performance. This is not surprising since the Scikit-learn implementation of classical k-means does not always produce clusters of equal size, and the quantum approach is running on imperfect hardware. The relative drop in performance between the classical and quantum approach is particularly apparent when $k = 2$. We suspect that the classical algorithms perform better for small values of $k$ because the number of ways to partition a data set increases dramatically as $k$ increases. When there are less possible ways to cluster a data set, a local solution to the training problem is more likely to be the correct partitioning of the data set. 

These experiments also show that the performance of our quantum algorithm degrades as problem size increases. We believe this is a reflection of the hardware that solved the QUBO problem rather than a flaw in our approach. The strong performance of the quantum algorithm on problems of size (8, 4), (12, 3), and (12, 4) give hope that as the fidelity and scale of quantum computers improves, our quantum approach may outperform its classical alternatives.

\begin{table*}[t!]
\centering
\caption{Time required to perform classical $k$-means, classical balanced $k$-means, and our QUBO formulation on datasets of increasing size. We also report approximate embedding time for the quantum approach.}
\label{tab:scalabilityN}

\begin{tabular}{c|cccc}
\hline\noalign{\smallskip}
\shortstack{Number \\ of points} & \shortstack{Classical \\ $k$-means} & \shortstack{Classical balanced \\
 $k$-means} & \shortstack{QUBO \\ formulation} & \shortstack{Embedding \\ (estimated)} \\
\noalign{\smallskip}\hline\noalign{\smallskip}
64 & 0.0218 $\pm$ 0.0017 & 0.0028 $\pm$ 0.0008 & 0.0008 $\pm$ 0.0003 & 0.1252 \\
128 & 0.0256 $\pm$ 0.0022 & 0.0073 $\pm$ 0.0025 & 0.0070 $\pm$ 0.0008 & 0.4973 \\
256 & 0.0334 $\pm$ 0.0035 & 0.0315 $\pm$ 0.0143 & 0.0192 $\pm$ 0.0018 & 1.9833 \\
512 & 0.0414 $\pm$ 0.0060 & 0.1637 $\pm$ 0.0607 & 0.1154 $\pm$ 0.0024 & 7.9224 \\ 
1024 & 0.0521 $\pm$ 0.0085 & 1.5577 $\pm$ 1.0501 & 0.4624 $\pm$ 0.0095 & 31.6696 \\
2048 & 0.0684 $\pm$ 0.0134 & 10.8928 $\pm$ 5.5405 & 1.8409 $\pm$ 0.0201 & 126.6392 \\
4096 & 0.1006 $\pm$ 0.0231 & 95.4876 $\pm$ 58.0103 & 7.6902 $\pm$ 0.0581 & 506.4798 \\
\noalign{\smallskip}\hline
\end{tabular}
\end{table*}

\begin{table*}[t!]
\centering
\caption{Time required to perform classical $k$-means, classical balanced $k$-means, and our QUBO formulation for datasets with an increasing number of clusters. We also report approximate embedding time for the quantum approach.}
\label{tab:scalabilityk}
\begin{tabular}{c|cccc}
\hline\noalign{\smallskip}
\shortstack{Number \\ of clusters} & \shortstack{Classical \\ $k$-means} & \shortstack{Classical balanced \\
 $k$-means} & \shortstack{QUBO \\ formulation} & \shortstack{Embedding \\ (estimated)} \\
\noalign{\smallskip}\hline\noalign{\smallskip}
2 & 0.02707	$\pm$ 0.0042 & 0.0276 $\pm$ 0.0101 & 0.0080 $\pm$ 0.0011 & 0.4973 \\
4 & 0.0427 $\pm$ 0.0058 & 0.0417 $\pm$ 0.0164 & 0.0198 $\pm$ 0.0021 & 1.9833 \\
8 & 0.0584 $\pm$ 0.0052 & 0.0399 $\pm$ 0.0113 & 0.1273 $\pm$ 0.0053 & 7.9224 \\
16 & 0.0873 $\pm$ 0.0089 & 0.0390 $\pm$ 0.0094 & 0.5129 $\pm$ 0.0271 & 31.6696 \\ 
32 & 0.1349 $\pm$ 0.0120 & 0.0271 $\pm$ 0.0052 & 1.9598 $\pm$ 0.0308 & 126.6392 \\
64 & 0.2341 $\pm$ 0.0090 & 0.0201 $\pm$ 0.0023 & 7.6511 $\pm$ 0.0695 & 506.4798 \\
\noalign{\smallskip}\hline
\end{tabular}
\end{table*}

\subsubsection{Scalability with Number of Data Points (N)} 
We also perform a scalability study to determine how the run time of our quantum approach varies as the number of data points increases. Due to the qubit limitations of modern adiabatic quantum computers, problems that require more than 64 binary variables ($N k > 64$) are impossible on the D-Wave 2000Q. However, we can approximate the run time of our algorithm on larger problems by measuring the time required to formulate the QUBO problem and the time required to postprocess a plausible solution. We estimate the time required to embed the problem ($t_e$) as well as annealing time ($t_a$). 

\begin{figure}[t!]
\centering
\includegraphics[scale = 0.6]{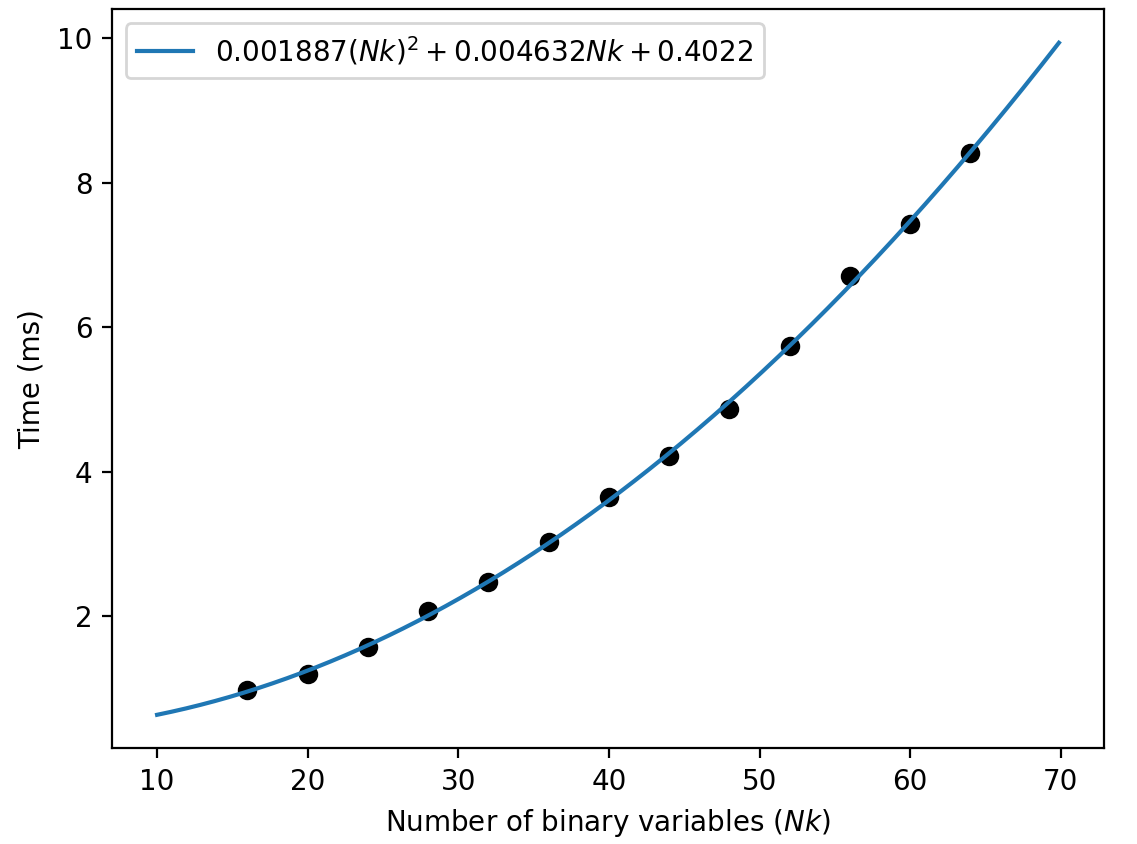}
\caption{Time required to embed small problems on the D-Wave using the embedding algorithm proposed by Date et. al. \cite{date2019efficiently}. Embedding time scales quadratically with the number of binary variables.}
\label{fig:embedding}     
\end{figure}

The runtime of the efficient embedding algorithm proposed in \cite{date2019efficiently} scales quadratically with the number of binary variables in the QUBO problem. 
Extrapolating upon the performance of this embedding algorithm on small problems, we approximate embedding time (in seconds) using the following equation:
\begin{align}
    t_e &= 1.887 \times 10^{-6} (Nk)^2 + 4.632 \times 10^{-6} (Nk) \nonumber \\
        & \qquad + 4.022 \times 10^{-4}
\end{align}

\begin{figure}[t!]
\centering
  \includegraphics[scale = 0.6]{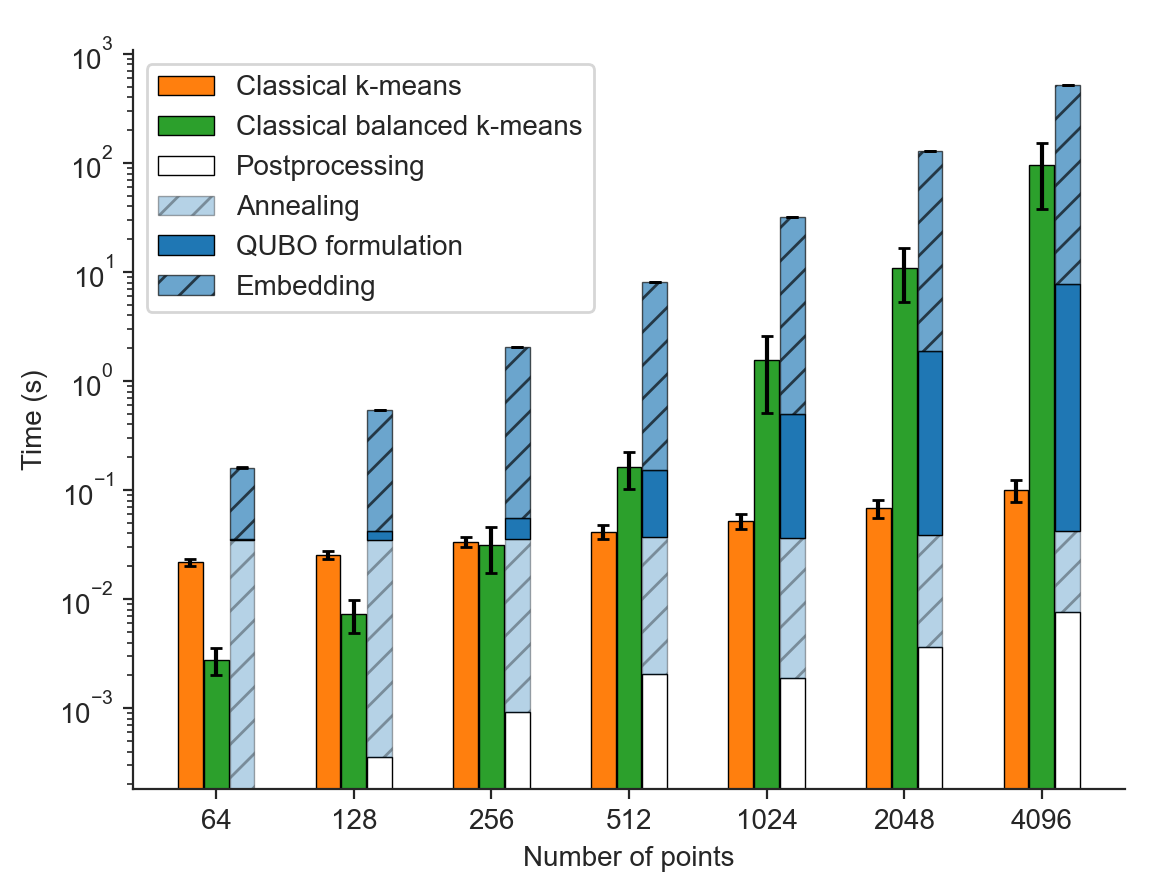}
\caption{Total computing time of classical $k$-means (orange bar), classical balanced $k$-means (green bar), and quantum balanced $k$-means (blue bar) as the number of points ($N$) in the training data set varies. Embedding and annealing times are approximate.}
\label{fig:scalabilityN}     
\end{figure}

As mentioned before, annealing is performed in constant time. Therefore, we assume that the annealing time for larger problems ($t_a$) is equal to the average annealing time for the small clustering experiments discussed in Section \ref{sec:synth}. 
\begin{align}
    t_a = 0.03481 \pm 0.00008
\end{align} 

\begin{figure}[t!]
\centering
  \includegraphics[scale = 0.6]{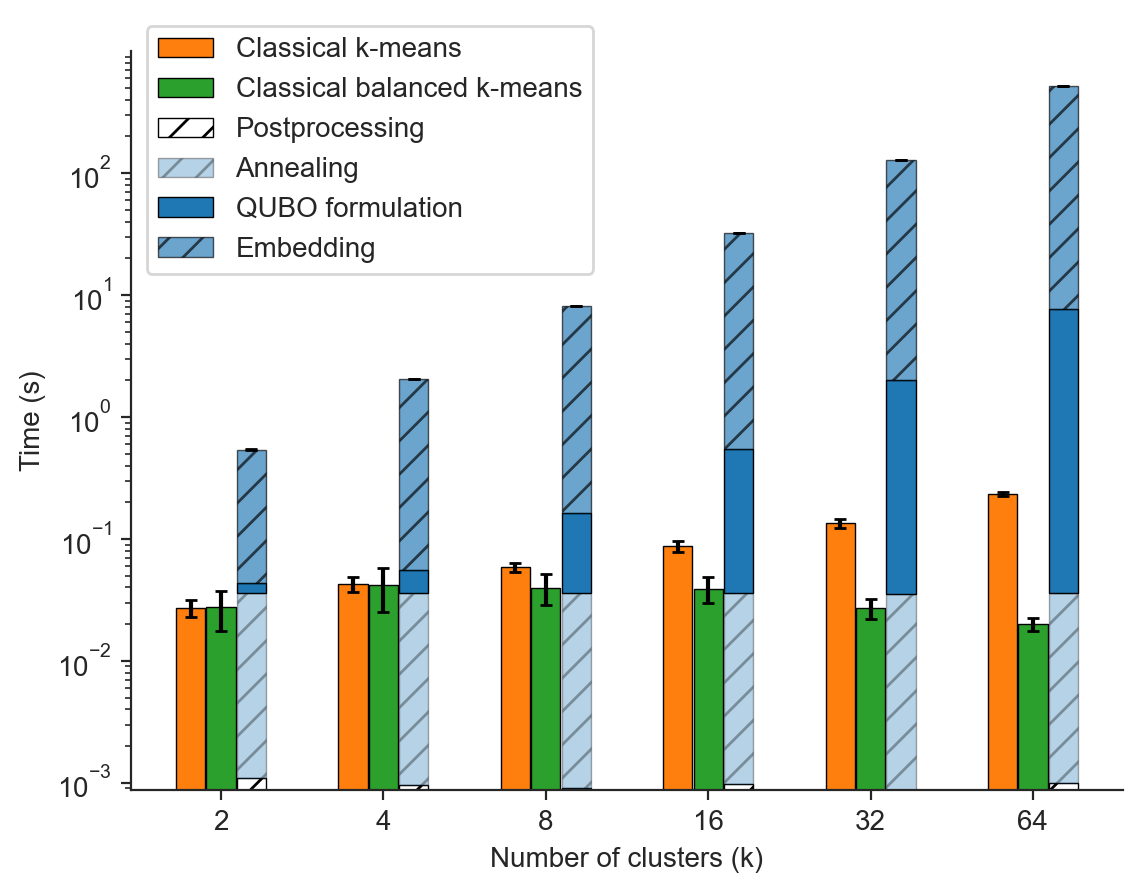}
\caption{Total computing time of classical $k$-means (orange bar), classical balanced $k$-means (green bar), and quantum balanced $k$-means (blue bar) as the number of clusters varies ($k$). Embedding and annealing times are approximate.}
\label{fig:scalabilityk}     
\end{figure}

We performed classical $k$-means, classical balanced $k$-means, and our QUBO formulation on data sets of increasing size. For a given problem type (defined by the number of points), all three approaches were run on 50 synthetic classification data sets. Each data set contained $k = 4$ classes, and each data point had $d = 2$ features. The average run time of each clustering approach is reported in Figure \ref{fig:scalabilityN} and Table \ref{tab:scalabilityN}.

\begin{table*}[t!]
\centering
\caption{Time required to perform classical $k$-means, classical balanced $k$-means, and our QUBO formulation on data sets with an increasing number of features. We also report approximate embedding time for the quantum approach.}
\label{tab:scalabilityd}
\begin{tabular}{c|cccc}
\hline\noalign{\smallskip}
\shortstack{Number \\ of features} & \shortstack{Classical \\ $k$-means} & \shortstack{Classical balanced \\
 $k$-means} & \shortstack{QUBO \\ formulation} & \shortstack{Embedding \\ (estimated)} \\
\noalign{\smallskip}\hline\noalign{\smallskip}
2 & 0.0508 $\pm$ 0.0089 & 1.5068 $\pm$ 0.6899 & 0.4742 $\pm$ 0.0185 & 31.6696 \\
4 & 0.0681 $\pm$ 0.0137 & 1.7589 $\pm$ 0.6546 & 0.4771 $\pm$ 0.0189 & 31.6696 \\
8 & 0.0803 $\pm$ 0.0105 & 2.2591 $\pm$ 1.0435 & 0.4737 $\pm$ 0.0101 & 31.6696 \\
16 & 0.4190 $\pm$ 0.1065 & 2.0672 $\pm$ 0.6473 & 0.4760 $\pm$ 0.0102 & 31.6696 \\ 
32 & 0.5411 $\pm$ 0.1171 & 2.1599 $\pm$ 0.6157 & 0.4895 $\pm$ 0.0179 & 31.6696 \\
64 & 0.6598 $\pm$ 0.1048 & 1.8983 $\pm$ 0.4888 & 0.5033 $\pm$ 0.0178 & 31.6696 \\
128 & 1.0369 $\pm$ 0.1577 & 1.6768 $\pm$ 0.4551 & 0.5283 $\pm$ 0.0221 & 31.6696 \\
256 & 1.2474 $\pm$ 0.1726 & 1.4060 $\pm$ 0.2184 & 0.5759 $\pm$ 0.0207 & 31.6696 \\
\noalign{\smallskip}\hline
\end{tabular}
\end{table*}

In each case, the quantum approach performed slower than both classical algorithms. However, the quantum run time was dominated by the embedding time. Embedding is extremely difficult on modern quantum computers due to limited qubit connectivity. As hardware improves, we expect embedding to be a considerably faster process. Therefore, on a future quantum computer, the quantum algorithm may outperform the classical balanced k-means algorithm for $N \geq 1024$, depending on how well the embedding process is optimized. Of the three approaches, our results indicate that the Scikit-learn implementation of classical $k$-means scales the best. This is expected since the time complexity of the Scikit-learn implementation of classical $k$-means ($\mathcal{O}(Nkd)$) is better than classical balanced $k$-means ($\mathcal{O}(N^3)$) or quantum balanced $k$-means ($\mathcal{O}(N^2kd)$).

\subsubsection{Scalability with Number of Clusters (k)}
\label{par:numclusters}
Following the same procedure, we analyze the scalability of each algorithm as the number of clusters $k$ is increased. For each problem type, all three clustering algorithms were run on 50 synthetic data sets. Each data set consisted of $N = 256$ points, and all points had $d = 8$ features. The average run time of each clustering approach is reported in Figure \ref{fig:scalabilityk} and Table \ref{tab:scalabilityk}.

In all cases, the quantum approach had a longer run time than both classical algorithms. Additionally, the quantum run time scaled worse as the number of clusters increased. This is expected since the third term in the QUBO formulation (Equation \ref{eq:kmeanscomplexity}) has time complexity $\mathcal{O}(Nk^2)$. Alternatively, classical $k$-means scales linearly with the number of clusters ($\mathcal{O}(Nkd)$), and balanced $k$-means clustering scales independently of the number of clusters ($\mathcal{O}(N^3)$). It is somewhat surprising that the average run time of the balanced $k$-means clustering approach decreases for $k > 16$. However, we suspect this is due to the smaller cluster sizes when $k$ is large.

\subsubsection{Scalability with Number of Features (d)}
\begin{figure}[h]
\centering
  \includegraphics[scale = 0.6]{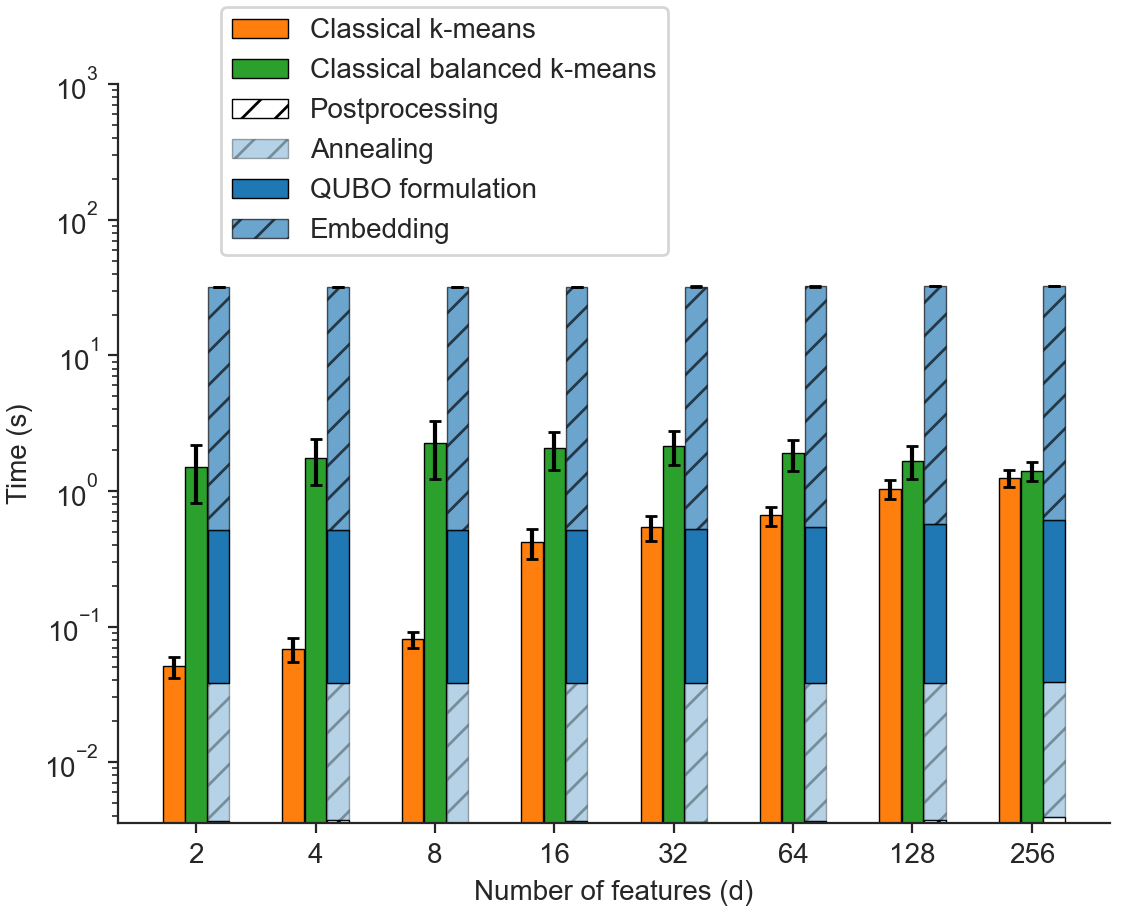}
\caption{We report the average total computing time of classical $k$-means (orange bar), classical balanced $k$-means (green bar), and quantum balanced $k$-means (blue bar) as the number of features ($d$) varies. Embedding and annealing times are approximate.}
\label{fig:scalabilityd}     
\end{figure}

Finally, we analyze the scalability of each algorithm with respect to the dimension of the training data set. For each problem type, all three clustering approaches were run on 50 synthetic classification data sets. Each data set consisted of $N = 1024$ points separated into $k = 4$ clusters. The average run time of each clustering approach is reported in Figure \ref{fig:scalabilityd} and Table \ref{tab:scalabilityd}.

As before, the quantum algorithm had the longest run time in all cases. However, on future hardware the quantum approach could perform better than classical $k$-means for $d \geq 128$ and better than classical balanced $k$-means for $d \leq 256$, depending on how well the embedding process is optimized. In Figure \ref{fig:scalabilityd}, it appears that the quantum approach scales better than classical $k$-means as $d$ increases. This is not surprising since the QUBO formulation only requires one computation related to the dimension of the data set (calculation of the distance matrix), while classical $k$-means requires distance calculations with each iteration. On the other hand, the quantum approach scales worse than classical balanced $k$-means. This is expected since the time complexity of classical balanced $k$-means is independent of $d$. It is somewhat surprising that the average run time of classical balanced $k$-means begins to decrease for $d > 32$, but we suspect this is due to cluster centers being farther apart on average.

\subsection{Clustering a Benchmark Data Set}

As a final proof of concept, we clustered portions of the Iris benchmark data set using our quantum clustering approach. This data set contains 150 points (each with 4 features) divided into 3 equal-size classes. Unfortunately, due to qubit limitations on modern hardware, it is impossible to perform quantum balanced k-means clustering on the entire data set. Therefore, we generate smaller data sets by picking $N / k$ points at random from $2 \leq k \leq 3$ of the data set's classes. For a given problem type (denoted by the number of points and number of clusters), all three clustering algorithms were run on 50 subsets of the Iris data set. Note that when $k=2$, all points were chosen from the first and second classes, which are linearly separable.

\begin{figure}[t!]
\centering
  \includegraphics[scale = 0.6]{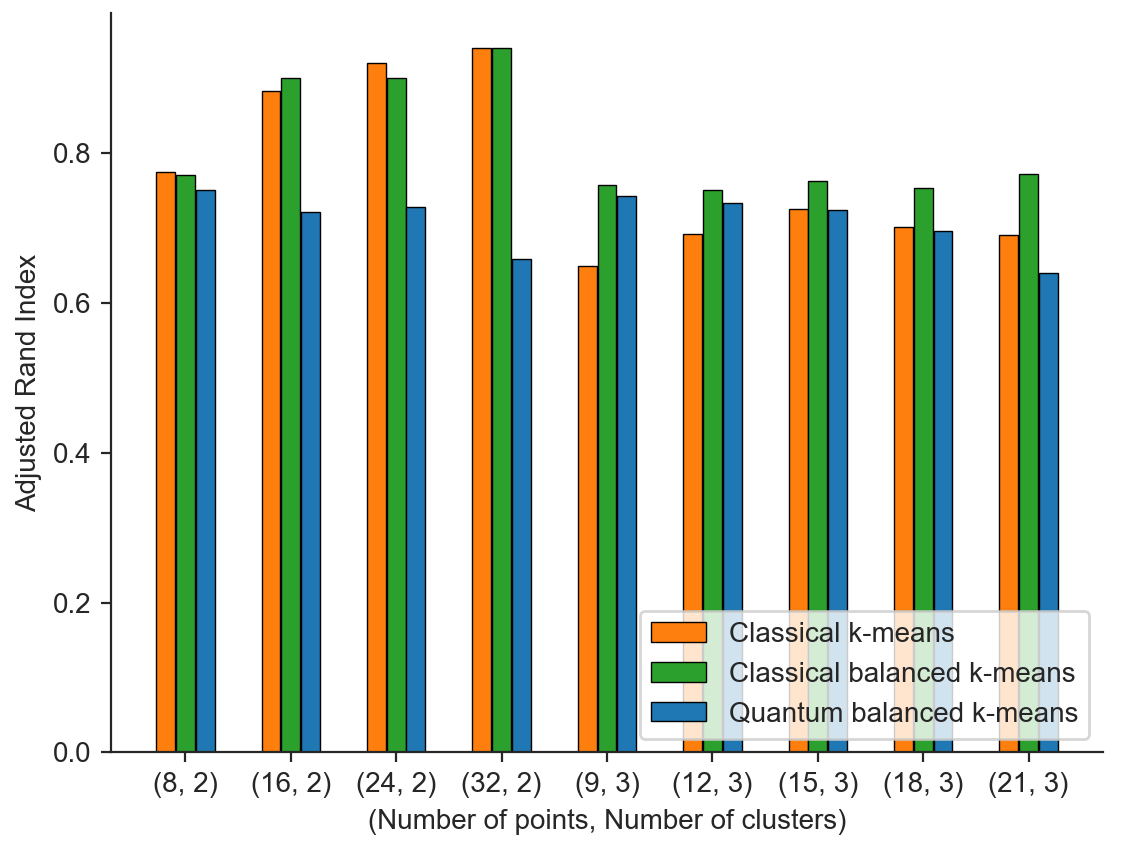}
\caption{Average adjusted rand index of classical $k$-means (orange bar), classical balanced $k$-means (green bar), and quantum balanced $k$-means (blue bar) on portions of the Iris data set.}
\label{fig:randiris}     
\end{figure}

For $k = 2$, the classical algorithms performed better than the quantum approach. This becomes particularly apparent as the number of binary variables ($Nk$) increases. For $k = 3$, the quantum algorithm has similar performance to classical balanced $k$-means and outperforms the Scikit-learn implementation of classical $k$-means for small data sets. Again, the performance of the quantum algorithm degrades as problem size increases. These results mirror the performance seen on the synthetic data sets discussed in Section \ref{sec:synth}.

\section{Conclusion}
\quad As new applications of machine learning models continue to emerge, it is of great interest to improve upon existing training algorithms. Adiabatic quantum computers are a promising alternative platform for solving NP-hard or NP-complete training problems efficiently. In this paper, we propose a quantum approach to training the balanced $k$-means clustering model. We analyze our approach theoretically, showing that it targets the global solution of the training problem better than its classical alternatives. We also show that our approach scales favorably on large data sets when compared to current classical balanced $k$-means algorithms. We test our approach using the D-Wave 2000Q adiabatic quantum computer and compare it to the Scikit-learn implementation of classical $k$-means as well as our own implementation of the classical balanced $k$-means algorithm with the best time complexity. We demonstrated that our quantum approach partitions data with similar accuracy to the classical approaches, even when running on imperfect hardware. As quantum hardware continues to improve in both fidelity and scale, we expect our approach to become a viable alternative to existing classical balanced clustering algorithms.

In the future, we hope to generalize our QUBO formulation to satisfy the generic $k$-means clustering training problem (Problem \ref{eq:kclusters1}). We also look to use elements of our approach to formulate quantum algorithms to similar clustering models, such as $k$-medoids clustering or fuzzy $C$-means clustering. Finally, we plan to investigate quantum approaches to clustering larger datasets within the qubit constraints of modern hardware.


%
\section*{Conflict of interest}
\quad The authors declare that they have no conflict of interest.

\bibliographystyle{IEEEtran}
\bibliography{references}

\end{document}